\documentclass[letterpaper, 10 pt, conference]{ieeeconf}  %

\IEEEoverridecommandlockouts                              %

\usepackage[utf8]{inputenc} %
\usepackage[T1]{fontenc}    %
\usepackage[hidelinks]{hyperref}       %
\usepackage{url}            %
\usepackage{booktabs}       %
\usepackage{amsfonts}       %
\usepackage{nicefrac}       %
\usepackage{microtype}      %
\usepackage[separate-uncertainty=true,detect-mode=true]{siunitx} %
\usepackage{mathtools}
\usepackage{amssymb}
\usepackage{amsmath}
\usepackage{tabu}
\usepackage{color}
\usepackage{algorithm}
\usepackage[noend]{algcompatible}

\usepackage{hyperref}
\usepackage[capitalize]{cleveref}
\usepackage{multirow}

\usepackage[style=ieee,backend=bibtex,url=false,doi=false,natbib=true,mincitenames=1,isbn=false]{biblatex}
 \AtEveryBibitem{\clearfield{issn}}
 \AtEveryCitekey{\clearfield{issn}}
 \usepackage[keeplastbox]{flushend}
 \usepackage{anyfontsize}

\DeclareMathOperator*{\E}{\mathbb{E}}

\newcommand{\paren}[1]{\mathopen{}\mathclose\bgroup\left(#1\aftergroup\egroup\right)}
\newcommand{\brock}[1]{\mathopen{}\mathclose\bgroup\left[#1\aftergroup\egroup\right]}
\newcommand{\curly}[1]{\mathopen{}\mathclose\bgroup\left\{#1\aftergroup\egroup\right\}}

\usepackage{amsmath,amssymb,mathtools}

\newcommand{\suchthat}{\;\ifnum\currentgrouptype=16 \middle\fi|\;}

\DeclarePairedDelimiterX{\infdivx}[2]{(}{)}{%
  #1\;\delimsize\|\;#2%
}

\title{\LARGE \bf
Online Parameter Estimation for Human Driver Behavior Prediction
}

\author{Raunak P. Bhattacharyya, Ransalu Senanayake, Kyle Brown, and Mykel J. Kochenderfer%
\thanks{R. P. Bhattacharyya, R. Senanayake, K. Brown, and M. J.  Kochenderfer are with the Stanford Intelligent Systems Laboratory in the Department of Aeronautics and Astronautics at Stanford University, Stanford, CA 94305, USA (email: \{raunakbh, ransalu, kjbrown7, mykel\}@stanford.edu).}
}

\begin{document}
\maketitle
\thispagestyle{empty}
\pagestyle{empty}

\begin{abstract}
Driver models are invaluable for planning in autonomous vehicles as well as validating their safety in simulation.
Highly parameterized black-box driver models are very expressive, and can capture nuanced behavior. However, they usually lack interpretability and sometimes exhibit unrealistic---even dangerous---behavior.
Rule-based models are interpretable, and can be designed to guarantee ``safe'' behavior, but are less expressive due to their low number of parameters.
In this article, we show that online parameter estimation applied to the Intelligent Driver Model captures nuanced individual driving behavior while providing collision free trajectories.
We solve the online parameter estimation problem using particle filtering, and benchmark performance against rule-based and black-box driver models on two real world driving data sets.
We evaluate the closeness of our driver model to ground truth data demonstration and also assess the safety of the resulting emergent driving behavior.
\end{abstract}
\section{Introduction} 
\label{sec:intro}
Modeling the behavior of human drivers is a key component in the development of safe automated vehicles.
    In the context of real-time behavior prediction on the road, the usefulness of a driver model depends on its ability to forecast the evolution of the traffic scene.
    Driver models are also important for safety validation in simulation.
        The effectiveness of simulation tests depends on generating a variety of scenarios that are representative of real human driving behavior~\cite{koopman2016challenges}.

Driver modeling is characterized by a high degree of uncertainty.
The behavior of any given vehicle depends on a multitude of unobservable psychological and physiological factors, including e.g., the driver's latent objectives and unique ``driving style.''
    Modeling is further complicated by interaction between multiple drivers.
        Even if all other sources of uncertainty in a traffic scene are ignored, this interaction between decision-making agents yields a complex multi-modal distribution over possible outcomes that can be very challenging to model.
\emph{Black-box} models (e.g., Gaussian mixture models, neural networks \cite{lefevre2014comparison}, \cite{morton2017analysis}, \cite{kuefler2017imitating}, \cite{bhattacharyya2019simulating}) often have the expressive power to capture nuanced driving behavior.
    However, such models lack interpretability and often exhibit unrealistic, even dangerous behavior (e.g., colliding with other vehicles) in regions of the state-space that are under-represented in the training dataset.
Though usually less expressive than black-box models, \emph{rule-based} models (e.g., the Intelligent Driver Model \cite{treiber2000congested}) are interpretable and---in many cases---can guarantee ``good behavior'' (e.g., collision-free driving).
    This ``good behavior'' arises directly from the model structure itself, which is informed by expert knowledge and applies even in regions of the state space that are under-represented in the data.

Model parameters can be selected \emph{offline} or \emph{online}.
    Offline methods can make use of arbitrary amounts of data because the duration of training is unconstrained. These methods usually yield ``average'' parameters for the population of drivers represented in the data set. Offline estimation is the paradigm of choice for essentially all black-box models and many rule-based models.
    In contrast, online methods can capture idiosyncrasies of individual drivers because these methods use real-time sensor information to select and/or update model parameters.
    The time and information implications of near real-time operation mean that online methods are best-suited to models (i.e., rule-based models) with relatively few parameters.

\begin{figure}[t]
    \centering
    \includegraphics[width=\columnwidth]{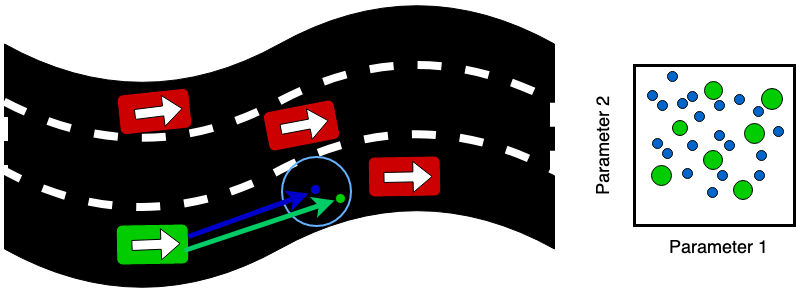}
    \caption{\footnotesize We model the behavior of a vehicle of interest (green) in the presence of other vehicles (red). How the green vehicle maneuvers is modeled using a stochastic extension of the Intelligent Driver Model. The parameters of the driver model are sequentially estimated from driving demonstration data using a particle filter. The bucket of particles is shown on the right. The new positions are sampled using blue particles and compared against green ground truth to weight the green particles.
    }
    \label{fig:motivation}
\end{figure}

In this paper, we apply online parameter estimation to an extension of the Intelligent Driver Model (IDM)~\cite{treiber2000congested} that explicitly models stochasticity in the behavior of individual drivers. 
Our specific contributions are:
\begin{itemize}
    \item We formulate a Bayesian estimation problem and propose a particle filtering approach for recursively estimating the parameters of the extended IDM (see~\cref{fig:motivation}).
    \item We use our parameter estimates for forward motion prediction. Model-generated trajectories are compared to ground truth trajectories from two real world driving data sets. Our model is benchmarked against various existing models. We also report emergent metrics for our model and the benchmark models.
\end{itemize}
The principles underlying our model are applicable in the general driver modeling problem.
However, in this paper we study the specific problem of \emph{car-following}, which reasons about longitudinal driving behavior.
\section{Background} 
\label{sec:background}
\subsection{Intelligent Driver Model and Extensions}
The IDM~\cite{treiber2000congested} is a parametric rule-based car-following model that balances two forces, the desire to achieve free speed if there were no vehicle in front, and the need to maintain safe separation with the vehicle in front.
The IDM is guaranteed to be collision free.
The inputs to the model are the vehicle's current speed $v(t)$ at time $t$, relative speed $r(t)$ with respect to the leading vehicle, and distance headway $d(t)$.
The model then outputs an acceleration according to
\begin{equation}
   a_{\mathrm{IDM}} = a_{\mathrm{max}}\Bigg( 1-\bigg(\frac{v(t)}{v_{\mathrm{des}}} \bigg)^4 - \bigg( \frac{d_{\mathrm{des}}}{d(t)} \bigg)^2 \Bigg) \text{,}
   \label{eqn:a_idm}
\end{equation}
where the desired distance is
\begin{equation}
    d_{\mathrm{des}} = d_{\mathrm{min}} + \tau .v(t) - \frac{v(t).r(t)}{2\sqrt{a_{\mathrm{max}}.b_{\mathrm{pref}}}} \text{.}
    \label{eqn:d_des}
\end{equation}

The model has several parameters that determine the acceleration output based on the scene information. Here, $v_{\mathrm{des}}$ refers to the free speed velocity, $d_{\mathrm{min}}$ refers to the minimum allowable separation between the ego and leader vehicle, $\tau$ refers to the minimum time separation allowable between ego and leader vehicle, $a_{\mathrm{max}}$ and $b_{\mathrm{pref}}$ refer to the limits on the acceleration and deceleration, respectively. Though the collision-free motion of a vehicle can be simulated by arbitrarily setting some parameter values, the driving behavior is not necessarily realistic. Therefore, in this paper, we learn the parameters from real human driver demonstrations.

\subsection{Related Work}
The literature contains various extensions of the original IDM.
The Enhanced IDM incorporates a slight modification that prevents the model from ``over-reacting'' when another vehicle cuts in front of it \cite{Kesting2010}.
The Foresighted Driver Model modifies the output of the IDM based on factors like upcoming curvature in the road \cite{Eggert}.
\citet{Liebner} incorporate a spatially varying velocity profile within the IDM to account for variation in different types of maneuvers through intersections.
    \citet{Hoermann2017} use a stochastic IDM model with fixed-variance additive Gaussian white noise. %
    \citeauthor{Schulz} use a similar model that also incorporates context-dependent upper and lower bounds on acceleration \cite{Schulz,Schulz2018}. 
Many approaches in the literature estimate IDM model parameters offline.
    \citet{lefevre2014comparison} use constrained nonlinear optimization.
    \citet{morton2017analysis} use the Levenberg-Marquardt algorithm.
    Some approaches select the parameters heuristically \cite{Schulz,Schulz2018}.
    In fact, ``recommended'' parameter values have been published for the IDM \cite{treiber2017intelligent}.

Offline estimation is also used for selecting parameter values in black-box driver models.
    \citet{lefevre2014comparison} use Expectation Maximization (EM) to train a Gaussian mixture model (GMM), and the Levenberg-Marquardt algorithm to train a neural network (NN).
    \citet{morton2017analysis} use gradient-based optimization to train various feedforward and recurrent neural network models.
    \citet{kuefler2017imitating} use Generative Adversarial Imitation Learning (GAIL) to train a recurrent neural network.

Some approaches estimate driver model parameters online.
    In Multi-Policy Decision-Making, the parameters of several hand-crafted control policies are estimated online with Bayesian Changepoint Estimation and Maximum-likelihood estimation \cite{Galceran2017a}.
    \citet{Sadigh2018} use online active information gathering to estimate the parameters of a human driver's reward function.

Several online estimation approaches are used for IDM in particular.
    \citet{monteil2015real} use an Extended Kalman filter.
    Examples of particle filters used with IDM parameters include approximate online POMDP solvers \cite{Sunberg2017a} and fully probabilistic scene prediction algorithms \cite{Hoermann2017}. The online parameter estimation approach of \citeauthor{Buyer2019} is similar to ours, although they use a different IDM extension and do not use their model for forward simulation of traffic scenes \cite{Buyer2019}. 
    None of the above models explicitly estimate ``stochasticity'' parameters for individual drivers.
        
\section{Method} 
\label{sec:method}
\subsection{Stochastic IDM}
To model human driving behavior, which is inherently stochastic (given the exact same scene, a human driver may not always take the same resulting action), we use the IDM with stochasticity~\cite{treiber2017intelligent}. 
We assume that the output acceleration is distributed according to 
\begin{equation}
    a \sim \mathcal{N}(a\mid a_{\mathrm{IDM}},\sigma_{\mathrm{IDM}}) \text{,}
    \label{eqn:errorable_idm}
\end{equation}
where $a_{\mathrm{IDM}}$ and $\sigma_{\mathrm{IDM}}$ represent the mean and variance, respectively, of a Gaussian distribution. The mean $a_{\mathrm{IDM}}$ is computed with \cref{eqn:a_idm}, and $\sigma_{\mathrm{IDM}}$ is a new model parameter.
Assuming the vehicle dynamics
\begin{equation}
    x_{t+1} = x_t + \frac{1}{2}a \Delta t^2 \text{,}
    \label{eqn:dynamics}
\end{equation}
\noindent where $x$ is the position and $\Delta t$ is the unit-time, we obtain the new position distributed according to
\begin{equation}
    x_{t+1} \sim \mathcal{N} (x_{t+1}\mid x_t + \frac{1}{2}a_{\mathrm{IDM}} \Delta t^2,\sigma_{\mathrm{IDM}} \Delta t^2) \text{.}
    \label{eqn:position_update}
\end{equation}

\subsection{Parameter Estimation}
\label{sec:paraest}
\begin{figure}[b]
    \centering
    \includegraphics[width=\columnwidth]{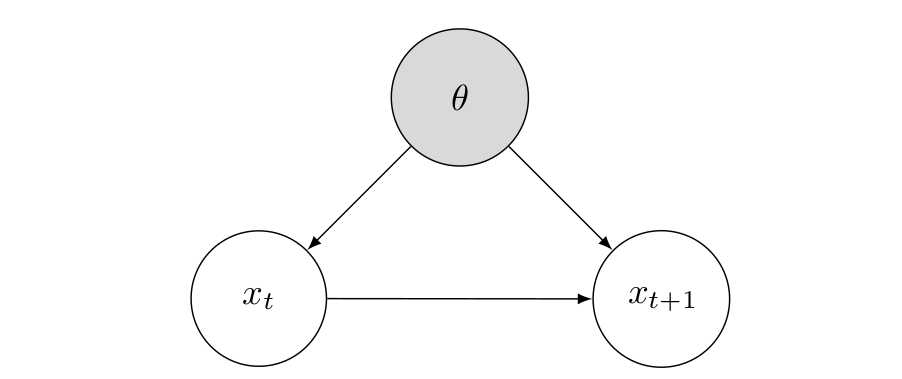}
    \caption{\footnotesize Bayesian network showing the driver modeling problem in the state estimation framework. $\theta $ represents the IDM parameters, $x_t$ and $x_{t+1}$ are the positions at timestep $t$ and $t+1$, respectively. The objective is to learn the latent IDM parameters $\theta$ from data collected from a human driver.
    }
    \label{fig:bayes_net}
    \setlength{\belowcaptionskip}{-15pt}
\end{figure}

We wish to maintain a distribution over model parameters for each driver in a given traffic scene. 
 We assume that the parameters $\theta_i$ of driver $i$ are stationary, i.e., human drivers do not change their latent driving behavior over the time horizons being considered in this work. 
While this assumption may not be entirely realistic (humans may have to change driving style given extreme situations), the relaxation to non-stationary parameters has been left to future work.
~\cref{fig:bayes_net} shows a Bayesian network where the hidden state contains the parameters of the IDM and the observations are the position trace obtained from demonstration data.

The distribution over parameters can be written as
\begin{equation}
    p(\theta \mid x_1,x_2,...,x_T) \text{,}
    \label{eqn:inference}
\end{equation}
\noindent where ($x_1,x_2,...,x_T$) denotes a sequence of position observations from driving demonstration data. This inference problem can be solved using recursive Bayesian estimation, where the recursive update equation is given by
\begin{equation}
    p(\theta \mid x_{1:t}) = \frac{p(x_t \mid \theta)p(\theta \mid x_{1:t-1})}
    {\int_{\theta} p(x_t \mid \theta)p(\theta \mid x_{1:t-1}) \mathrm{d}\theta} \text{.}
    \label{eqn:recursive_state_estimation}
\end{equation}

The partition function (the denominator) in~\cref{eqn:recursive_state_estimation} cannot be evaluated analytically for general nonlinear distributions.
Rather than imposing restrictive assumptions on the form of the distribution, we use particle filtering~\cite{thrun2002particle,Thrun:2005:PR:1121596} to approximately solve the inference problem.
A particle filter approximates a continuous probability distribution with a collection of samples, called particles.

Given a starting scene, each vehicle has an associated initial set of $I$ particles contained in the set $\Theta = \{ \theta_1, \theta_2, \dots, \theta_I\}$
For all these vehicles, demonstration trajectories of length $T$ are given.
At every time-step, a random particle $\theta_i$ is selected from the set $\Theta$ and a new position $x_i^{(t+1)}$ for the vehicle is sampled from the generative model defined by the IDM with parameters represented by the particle.
We explicitly capture the interaction between vehicles by sampling the new position in the presence of other vehicles driven by a deterministic IDM.
The particle is then weighted according to the likelihood of the true position $x^{(t+1)}$ under the distribution given by the sampled position $x_i^{(t+1)}$.
This likelihood is computed by querying the Gaussian probability density function with mean given by $x_i^{(t+1)}$ and variance given by $\sigma_\mathrm{IDM} \Delta t^2$ per~\cref{eqn:position_update}.
Finally, particles are resampled according to the weights to yield a new set of particles. The particle filtering algorithm is given in~\cref{algo:pfidm}. 
\begin{algorithm*}
  \caption{IDM parameter estimation using particle filtering}
  \label{algo:pfidm}
  \begin{algorithmic}
    \STATE {\bfseries Input:} Expert trajectories of length $T$, Starting scene with $K$ vehicles, Initial set of particle sets $\{\Theta_1, \Theta_2, \dotsc, \Theta_K\}$
    \FOR[time-steps]{$t \gets 0, 1, \dotsc, T$}
    \FOR[vehicles] {$k \gets 1,2, \dotsc, K$}
    \STATE $x_k^{(t+1)} \gets$ true position of the $k$th vehicle at $t+1$
    \FOR[particles]{$i \gets 1, 2, \dotsc, I_k$}
    \STATE $\theta_i \gets$ random particle in $\Theta_k$
    \STATE $x_{k,i}^{(t+1)} \sim \mathcal{N}( x_k^{(t)}+\frac{1}{2}a_{\mathrm{IDM}_{\theta_i}}\Delta{t}^2,\sigma_{\mathrm{IDM}_{\theta_i}}\Delta{t}^2)$ \COMMENT{Sampled next position of the $k$th vehicle using the $i$th particle - \cref{eqn:position_update}}
    \STATE $w_i \gets O\big( x_k^{(t+1)}\mid x_{k,i}^{(t+1)} \big)$ \COMMENT{Probability density of true next position given sampled next position}
    \ENDFOR
    \STATE $\Theta_k \gets$ Obtain $I_k$ samples from $\Theta_k$ according to $[w_1, w_2, \dotsc, w_{I_k}]$ \COMMENT{Resampling}
    \ENDFOR
    \ENDFOR
  \end{algorithmic}
\end{algorithm*}

Particle filtering is susceptible to the particle deprivation problem wherein particles converge to one region of the state space and there is no exploration of other regions. 
Dithering~\cite{schon2011particle} helps to mitigate the particle deprivation problem wherein external noise is added to aid exploration of state space regions.
We implement dithering by adding random noise to the top 20\% particles ranked according to the corresponding likelihood.

Rather than estimating all model parameters, we estimate only the desired velocity ($v_{\mathrm{des}}$) and the driver-dependent stochasticity ($\sigma_{\mathrm{IDM}}$) as the parameters to be estimated. 
Therefore, the parameter space is two-dimensional, with $\theta=[v_{\mathrm{des}},\sigma_{\mathrm{IDM}}]$.

At the end of the particle filtering process, every vehicle has an associated collection of particles that should best explain the observed driving behavior given in the demonstration trajectory.
IDM parameters are then extracted from this collection of particles to drive the vehicles in simulation and collect metrics to assess the driving behavior.
\section{Experiments} 
\label{sec:experiments}
We evaluate the performance of our model on demonstration data from two real world datasets, namely the Next-Generation Simulation (NGSIM) for US Highway 101~\cite{colyar2007us} which provides driving data collected at \SI{10}{\hertz} and the Highway Drone Dataset (HighD)~\cite{highDdataset} which provides driving data from German highways recorded at \SI{25}{\hertz} using a drone.
    We benchmark our approach against representative rule-based and black-box models as well as constant velocity and constant acceleration baselines.

Experiments are conducted on a set of thirty scenarios (fifteen scenarios randomly sampled from each dataset). Twenty vehicles in each scenario are randomly selected as target vehicles.
For each scenario and each model, predicted trajectories are generated by forward simulation of this set of target vehicles over a \SI{5}{\second} time horizon, where the target vehicles are controlled by the driver model defined by the parameters estimated using particle filtering.

We use Root Mean Squared Error (RMSE) of the position and velocity to measure ``closeness'' of a predicted trajectory to the corresponding ground-truth trajectory.
\Cref{fig:rmse} shows the RMSE over time for an example scenario with 20 vehicles over a \SI{5}{\second} duration from the NGSIM dataset.

While RMSE measures prediction accuracy at the level of individual vehicles by comparing the obtained trajectories against ground truth from the demonstration trajectories, we also wish to quantify how ``safely'' each model drives. To this end, we count the number of ``undesirable events'' (collision, going off the road, and hard braking) that occur in each scene prediction.

The code for all the experiments is publicly available at our code base.\footnote{\href{https://github.com/sisl/ngsim_env/tree/idm_pf_NGSIM}{https://github.com/sisl/ngsim\_env/tree/idm\_pf\_NGSIM}}

\subsection{Filtering}
To estimate the parameters of the IDM using our filtering approach as per~\cref{algo:pfidm}, the particles are initially sampled from a uniform distribution discretized into a grid with resolution of \SI{0.5}{\meter/\second} for the desired velocity parameter ($v_{\mathrm{des}}$) and $0.1$ for the stochasticity parameter ($\sigma_{\mathrm{idm}}$). 
At the dithering stage (to avoid particle deprivation), we add noise sampled from a discrete uniform distribution with $v_{\mathrm{des}} \in \{-0.5,0,0.5\}$ and $\sigma_{\mathrm{IDM}} \in \{-0.1,0,0.1\}$.
These values are chosen to preserve the discretization present in the initial sampling of particles.
The time taken for filtering to converge in a 20 vehicle scenario over a \SI{5}{\second} duration was \SI{30}{\second} on an Intel Core i9-9900K eight-core processor.

\begin{figure}
    \centering
    \includegraphics[width=\columnwidth]{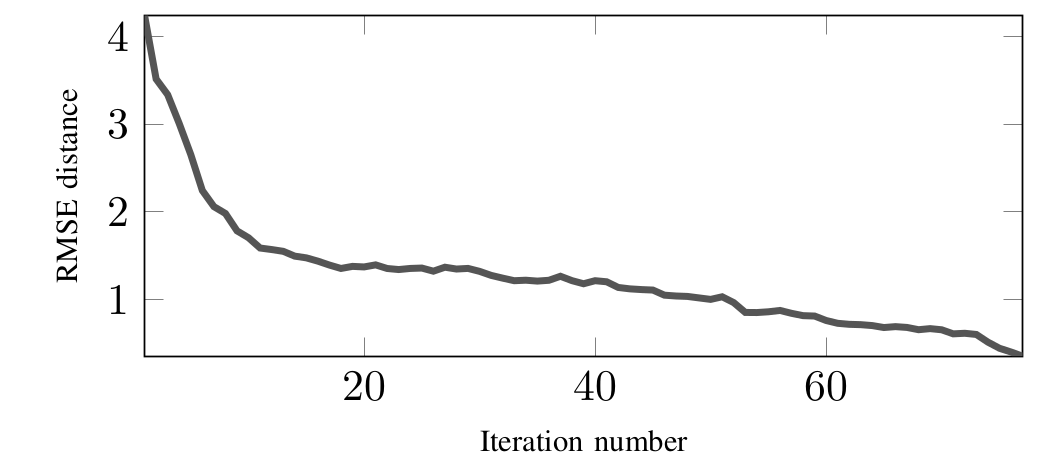}
    \caption{\footnotesize RMSE distance from final particle over particle set at every iteration averaged over all the vehicles. The particle set converges to the final particle with the progress of filtering.}
    \label{fig:filtering_progress}
    \setlength{\belowcaptionskip}{-15pt}
\end{figure}

To assess the convergence of the particle filtering approach,~\cref{fig:filtering_progress} shows the root mean squared distance from the mean of the final particle distribution over the set of particles at every iteration.
The particles converge as more demonstration data is shown to the filtering algorithm.

\begin{figure}
    \centering
    \includegraphics[width=\columnwidth]{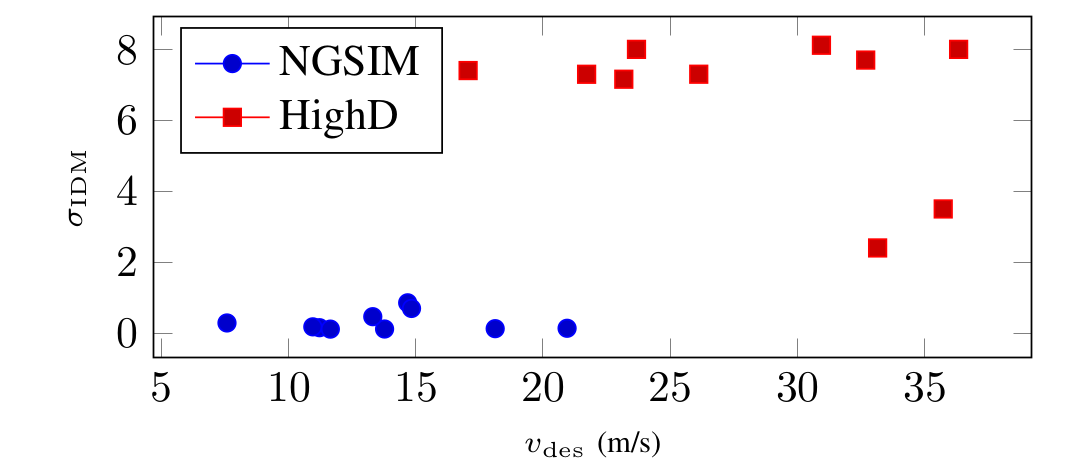}
    \caption{\footnotesize Mean particles from final distributions achieved after particle filtering for a set of 10 NGSIM and HighD vehicles observed over trajectories of 50 timesteps. HighD vehicles are faster on average.}
    \label{fig:carwise_mean_particle}
    \setlength{\belowcaptionskip}{-15pt}
\end{figure}

\Cref{fig:carwise_mean_particle} shows the mean particle after the filtering process for a subset of vehicles from both the NGSIM and the HighD datasets. 
The HighD vehicles have a higher desired velocity ($v_\mathrm{des}$) parameter on average, reflecting the fact that vehicles drive faster on German highways.

\subsection{Benchmarking}
To benchmark the performance of our approach, we compare the driving behavior obtained by our model against that obtained by five other models. %
The first benchmark model is IDM with the ``default'' parameter values recommended in ~\cite{treiber2017intelligent}: $v_{\mathrm{des}} = \SI{30}{\meter/\second}$, $\tau=\SI{1.0}{\second}$, $d_{\mathrm{min}}=\SI{2}{\meter}$, $a_{\mathrm{max}} = \SI{3}{\meter/\second^2}$, and $b_{\mathrm{pref}} = \SI{2}{\meter/\second^2}$.
Our second (also rule-based) benchmark model is the IDM with parameters obtained by offline estimation using non-linear least squares~\cite{morton2017analysis}.
The associated parameter values are $v_{\mathrm{des}} = \SI{17.837}{\meter/\second}$, $\tau= \SI{0.918}{\second}$, $d_{\mathrm{min}}=\SI{5.249}{\meter}$, $a_{\mathrm{max}} = \SI{0.758}{\meter/\second^2}$, and $b_{\mathrm{pref}} = \SI{3.811}{\meter/\second^2}$.
Our third benchmark model is a recurrent network trained with Generative Adversarial Imitation Learning (GAIL)~\cite{bhattacharyya2018multi}.
We also baseline our method against constant velocity (vehicles continue driving at the same speed that they start with at the beginning of the simulation) and constant acceleration (vehicles accelerating at $\SI{1}{\meter/\second^2}$) models.

\begin{figure}
  \centering

    \includegraphics[width=\columnwidth]{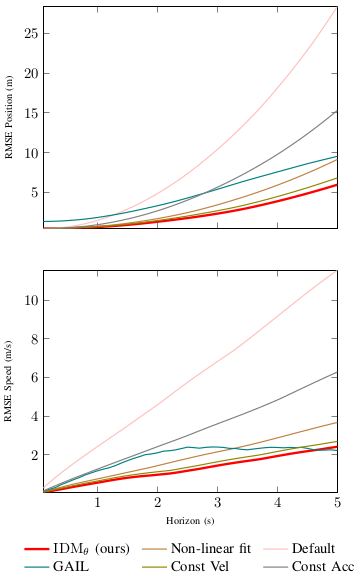}
    
  \setlength{\belowcaptionskip}{-15pt}
  \caption{\footnotesize
  Root mean square error in position and velocity averaged over all cars to benchmark our model ($\mathrm{IDM}_{\theta}$) against other driver models. Default refers to an IDM with parameters as set to default for motorways~\cite{treiber2017intelligent}. Non-linear fit refers to an IDM with parameters estimated offline from data using non-linear least-squares fit~\cite{morton2017analysis}. GAIL refers to a black box driver model trained using Generative Adversarial Imitation Learning~\cite{bhattacharyya2018multi}. Baseline models are constant acceleration and constant velocity driving models.
  }
  \label{fig:rmse}
\end{figure}

\subsection{Results}

RMSE results for an example scenario with 20 vehicles over a \SI{5}{\second} duration from the NGSIM dataset are shown in~\cref{fig:rmse}.
We observe that our method provides driving trajectories that are closer to the ground truth as compared to those generated by IDM with default parameter values, and those generated by GAIL driven policies.
    We see that the RMSE in both position and velocity averaged over the set of vehicles is lowest for all timesteps using our driving model.

Further experiments on both NGSIM and HighD datasets are reported in~\cref{table:comp}. These results are generated using 15 randomly sampled scenarios from both the HighD and NGSIM datasets. 
Every scenario is such that there is a set of 20 vehicles driving over a \SI{5}{\second} horizon which translates to 50 timesteps for NGSIM and 125 timesteps for HighD.
We see that while our method outperforms other methods, it performs worse than the constant velocity baseline for the HighD dataset. 
One possible reason may be the default values for the parameters that govern the interaction between vehicles, i.e. minimum allowed separation $d_\mathrm{min}$ and minimum timegap $\tau$.
Including these parameters within the filtering process will allow finer grained driver modeling and is an interesting direction for future work.

\begin{table*}
  \centering
    \caption{\footnotesize Experiments over 15 randomly selected scenarios for both NGSIM and HighD each with 20 vehicles driving for a $5$ \si{s} duration. The results show the RMSE values for position and velocity at the end of $5$ \si{s}. Cumulative number of collisions at the end of the horizon are also reported.}
    \begin{tabular}{@{}llcccccc@{}}
    \toprule
     &&\multicolumn{5}{c}{\bf Models}  \\ \cmidrule{3-8}
      {\bf Metrics}&{\bf Dataset} & $\mathrm{IDM}_{\theta}$ (ours) & Default \cite{treiber2017intelligent} & GAIL \cite{bhattacharyya2018multi} & Const. Speed & Const. Acc. & Non-Linear Fit~\cite{morton2017analysis}\\
      \midrule
      \parbox[t]{2.2cm}{\multirow{2}{*}{Position RMSE}}  
      &NGSIM & 5.90 $\pm$ 1.98 &
      27.78 $\pm$ 5.40 &
      10.42 $\pm$ 3.73 & 
      6.24 $\pm$ 2.02 &
      12.64 $\pm$ 4.70 &
      7.34 $\pm$ 4.55\\
     &HighD & 8.02 $\pm$ 3.34 &
     18.30 $\pm$ 9.03 & 
     13.63 $\pm$ 3.92 &
     2.42 $\pm$ 1.64 & 
     11.01 $\pm$ 1.92 &
     35.13 $\pm$ 7.21\\
      \midrule
     \parbox[t]{2.2cm}{\multirow{2}{*}{Velocity RMSE}}
     &NGSIM & 2.12 $\pm$ 0.79 &
     10.72 $\pm$ 2.36 &
     3.52 $\pm$ 1.28 &
     2.22 $\pm$ 0.82 &
     5.03 $\pm$ 1.78 &
     2.69 $\pm$ 1.77\\
       &HighD & 2.14 $\pm$ 0.65 &
       4.59 $\pm$ 2.46 &
       2.94 $\pm$ 0.93 &
       0.94 $\pm$ 0.57 &
       4.39 $\pm$ 0.61 &
       10.05 $\pm$ 2.07\\
      \midrule
      \parbox[t]{2.2cm}{\multirow{2}{*}{Number of collisions}}
      &NGSIM & 0 $\pm$ 0 &
      0 $\pm$ 0 &
      53 $\pm$ 11 &
      113 $\pm$ 18 &
      119 $\pm$ 16 &
      0 $\pm$ 0\\
       &HighD & 0 $\pm$ 0 &
       0 $\pm$ 0 & 
       15 $\pm$ 4 &
       0 $\pm$ 0 &
       27 $\pm$ 3 &
       0 $\pm$ 0\\
      \bottomrule
    \end{tabular}
  \label{table:comp}
\end{table*}

\begin{figure}
    \centering
    \includegraphics[width=\columnwidth]{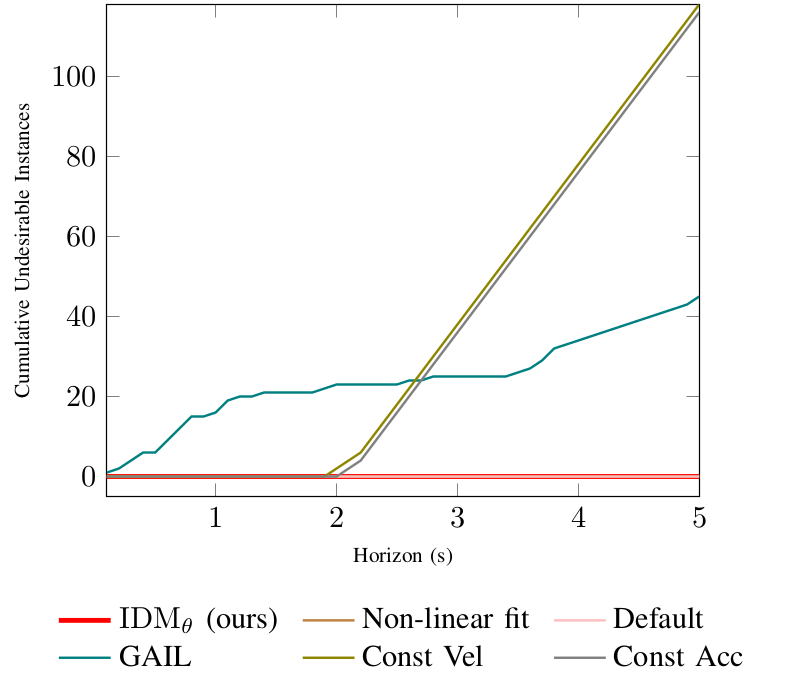}
    \caption{\footnotesize Cumulative number of undesirable instances summed over all vehicles over a 5 \si{s} time horizon using different driving models in a congested scenario from the NGSIM dataset. IDM based models, including ours, result in no collisions, off-the-road driving, or hard decelerations.
    }
    \setlength{\belowcaptionskip}{-15pt}
    \label{fig:undes_cumsum}
\end{figure}

The cumulative number of undesirable driving instances for 20 vehicles over a \SI{5}{\second} duration in a congested traffic scenario from the NGSIM dataset is shown in ~\cref{fig:undes_cumsum}.
    The cumulative number of undesirable driving instances keep growing with time for the data-driven benchmark in~\cref{fig:undes_cumsum}.
This reflects the fact that GAIL does not provide guarantees on safety.
As expected, the IDM based models including ours, and the two rule-based benchmarks do not show any collisions because the IDM is collision-free by default.
The constant velocity and constant acceleration baselines also do not provide collision-free trajectories because they are not reacting to the vehicle in front of them but merely driving with constant velocity and acceleration, respectively.

Cumulative number of collisions for all vehicles over the duration of the trajectory are also reported in~\cref{table:comp}.
We observe that the constant velocity baseline suffers from no collisions in the HighD dataset.
This is because the dataset is not as congested as the NGSIM dataset and hence vehicles start with sufficient distance headway and relative velocity to avoid collisions.
However, the constant acceleration does result in some collisions whenever a faster vehicle starts out behind a slower vehicle.
The data-driven benchmark also results in some collisions (fewer than NGSIM due to larger separation between vehicles).
As expected, congested scenarios present a challenge for the benchmark models.

\section{Conclusions} 
\label{sec:conclusions}
In this paper, we proposed a methodology that learned the parameters of the stochastic Intelligent Driver Model from driving demonstration data.
We used particle filtering to perform online estimation of the parameters of the stochastic IDM.
We benchmarked our driving model on two real driving datasets against both rule-based models and black-box driving models.
We assessed the driving performance both in terms of closeness to demonstration trajectories as well as safety of emergent driving behavior.

While we assumed fixed parameters of the Intelligent Driver Model, future work will investigate the impact of changing parameters to account for nonstationarity in human driver behavior. 
While this work demonstrated the proposed approach only on two parameters, future work will also extend the particle filtering method to more parameters leading to a finer grained characterization of driving behavior.
The driving model will also be combined with a lane changing model such as MOBIL~\cite{kesting2007general} to extend to two-dimensional driving behavior.
Datasets involving driving behavior with labeled driving style will be used to assess the capability of the proposed methodology to capture individual driving behavior from demonstrations.
Finally, the resulting driving models will be used for generating reliable simulations of human driving to enable safety assessment of autonomous vehicles.

\section*{Acknowledgments}
Toyota Research Institute (TRI) provided funds to assist the authors with their research, but this article solely reflects the opinions and conclusions of its authors and not TRI or any other Toyota entity. The authors thank Jeremy Morton and Louis Dressel for useful discussions.

\renewcommand*{\bibfont}{\footnotesize}
\printbibliography

\end{document}